\documentclass[sigconf]{acmart}

\AtBeginDocument{%
  }

\copyrightyear{2025}
\acmYear{2025}
\setcopyright{acmlicensed}
\acmConference[CIKM '25]{Proceedings of the 34th ACM International Conference on Information and Knowledge Management}{ November 10--14, 2025}{Seoul, Republic of Korea}
\acmBooktitle{Proceedings of the 34th ACM International Conference on Information and Knowledge Management (CIKM '25), November 10--14, 2025, Seoul, Republic of Korea}
\acmDOI{10.1145/3746252.3760803}
\acmISBN{979-8-4007-2040-6/2025/11}
\received[revised]{August 2025}
\received[accepted]{by cikm 2025}

\usepackage{xcolor} 
\usepackage{multirow,enumitem}
\usepackage{wrapfig}
\usepackage{balance}
\usepackage{url}
\usepackage{hyperref}
\hypersetup{
    colorlinks=true,
    citecolor=blue,
    linkcolor=blue,
    filecolor=blue,
    urlcolor=magenta,
    unicode=true
    }

\begin{document}

\title{Integrating Time Series into LLMs via Multi-layer Steerable Embedding Fusion for Enhanced Forecasting}


\settopmatter{authorsperrow=4}
\author{Zhuomin Chen}
\orcid{0009-0002-4406-7947}
\affiliation{%
  \department{School of Software Engineering}
  \institution{Sun Yat-Sen University}
  \city{Zhuhai}
  \state{Guangdong}
  \postcode{519080}
  \country{China}
}
\email{chenzhm39@mail2.sysu.edu.cn}

\author{Dan Li}
\authornote{Corresponding author.}
\orcid{0000-0002-3787-1673}
\affiliation{%
  \department{School of Software Engineering}
  \institution{Sun Yat-Sen University}
  \city{Zhuhai}
  \state{Guangdong}
  \postcode{519080}
  \country{China}
}
\email{lidan263@mail.sysu.edu.cn}

\author{Jiahui Zhou}
\orcid{0009-0002-2464-4102}
\affiliation{%
  \department{School of Software Engineering}
  \institution{Sun Yat-Sen University}
  \city{Zhuhai}
  \state{Guangdong}
  \postcode{519080}
  \country{China}
}
\email{zhoujh99@mail2.sysu.edu.cn}

\author{Shunyu Wu}
\orcid{0009-0003-1921-2767}
\affiliation{%
  \department{School of Software Engineering}
  \institution{Sun Yat-Sen University}
  \city{Zhuhai}
  \state{Guangdong}
  \postcode{519080}
  \country{China}
}
\email{wushy88@mail2.sysu.edu.cn}

\author{Haozheng Ye}
\orcid{0009-0005-9927-8911}
\affiliation{%
  \department{School of Software Engineering}
  \institution{Sun Yat-Sen University}
  \city{Zhuhai}
  \state{Guangdong}
  \country{China}
}
\email{yehzh8@mail2.sysu.edu.cn}

\author{Jian Lou}
\orcid{0000-0002-4110-2068}
\affiliation{%
  \department{School of Software Engineering}
  \institution{Sun Yat-Sen University}
  \city{Zhuhai}
  \state{Guangdong}
  \postcode{519080}
  \country{China}
}
\email{louj5@mail.sysu.edu.cn}

\author{See-Kiong Ng}
\orcid{0000-0001-6565-7511}
\affiliation{%
  \department{Institute of Data Science}
  \postcode{117602}
  \department{School of Computing}
  \postcode{117417}
  \institution{\makebox[0pt][c]{National University of Singapore}}
  \city{Singapore}
  \country{Singapore}
}
\email{seekiong@nus.edu.sg}

\renewcommand{\shortauthors}{Zhuomin Chen et al.}

\begin{abstract}
Time series (TS) data are ubiquitous across various application areas, rendering time series forecasting (TSF) a fundamental task. With the astounding advances in large language models (LLMs), a variety of methods have been developed to adapt LLMs for time series forecasting. Despite unlocking the potential of LLMs in comprehending TS data, existing methods are inherently constrained by their shallow integration of TS information, wherein LLMs typically access TS representations at shallow layers, primarily at the input layer. This causes the influence of TS representations to progressively fade in deeper layers and eventually leads to ineffective adaptation between textual embeddings and TS representations. In this paper, we propose the Multi-layer Steerable Embedding Fusion (MSEF), a novel framework that enables LLMs to directly access time series patterns at all depths, thereby mitigating the progressive loss of TS information in deeper layers. Specifically, MSEF leverages off-the-shelf time series foundation models to extract semantically rich embeddings, which are fused with intermediate text representations across LLM layers via layer-specific steering vectors. These steering vectors are designed to continuously optimize the alignment between time series and textual modalities and facilitate a layer-specific adaptation mechanism that ensures efficient few-shot learning capabilities. Experimental results on seven benchmarks demonstrate significant performance improvements by MSEF compared with baselines, with an average reduction of 31.8\% in terms of MSE. The code is available at \textcolor{magenta}{\url{https://github.com/One1sAll/MSEF}}.
\end{abstract}



\maketitle

\section{Introduction}
Time series (TS) data is pervasive in various industrial sectors, such as finance \cite{sezer2020financial, zhang2025camef, wang2025pre}, meteorology \cite{fathi2022big}, transportation \cite{aouedi2025deep}, and energy systems \cite{mystakidis2024energy}. Accurately forecasting the future trend of TS data facilitates proactive resource management and risk mitigation strategies. 
Recent advances in large language models (LLMs) have inspired numerous efforts to adapt the remarkable pattern extrapolation capabilities of large-scale pretrained foundation models to time series forecasting (TSF) \cite{xue2023promptcast,liu2024lstprompt,jin2023time,Gruver2023large}.

Existing strategies adapt LLMs to TSF tasks are two-folded: (1) \textbf{textualization approaches} that manually transform numerical time series values into natural language descriptions through numerical digits-to-text/tokens conversion \cite{xue2023promptcast,Gruver2023large}, statistical feature extraction \cite{jin2023time,xue2023promptcast}, or external knowledge incorporation \cite{liu2024lstprompt, jiangfstllm, zhang2025camef}; and (2) \textbf{vectorization techniques} that employ encoding layers to project raw time series into continuous embeddings spaces which is made compatible with LLM input spaces \cite{jin2023time}. In addition, knowledge distillation methods, like TimeKD\cite{liu2025efficient}, focus on efficiency through privileged knowledge distillation to transfer LLM capabilities to lightweight student models. While these methods demonstrate initial success in enabling LLMs to process TS data, they share a fundamental limitation: adaptations are confined to the input level through either linguistic transformation or embedding projection, without addressing the core architecture of LLMs. Crucially, neither approach explicitly incorporates essential TS embeddings into the LLM's attention mechanisms nor hierarchical layer-wise representations, leading to suboptimal performance in downstream forecasting tasks.

This limitation stems from treating LLMs as static, language-oriented models that simply ``interpret'' time series data rather than dynamically adjusting their internal mechanisms (e.g., attention patterns, tokenization strategies, or representation hierarchies) to accommodate TS data. Consequently, a significant research gap remains: existing works have not comprehensively investigated how to deeply incorporate TS characteristics into LLMs' fundamental architectures beyond superficial input modifications.

In this work, we propose \textbf{Multi-layer Steerable Embedding Fusion (MSEF)}, a novel approach for few-shot time series forecasting that enables comprehensive integration of TS patterns across all layers of the adopted LLM. MSEF overcomes information dissipation in deeper layers by: (1) leveraging time series foundation models (TSFMs) to extract semantically meaningful TS embeddings, and (2) introducing specialized steering vectors that continuously optimize the alignment between TS and textual representations during forward propagation across the LLM layers. Furthermore, the layer-specific adaptation mechanism of MSEF ensures efficient few-shot learning while maintaining superior forecasting performance compared to conventional supervised approaches. Our key contributions can be summarized as follows.

\begin{itemize}
\item \textbf{Multi-layer Temporal Fusion}: By injecting time series representation at each LLM layer, MSEF maintains the integrity of temporal patterns throughout the LLM architecture, enabling hierarchical TS understanding while preventing information loss compared to single-layer adaptation methods. 
\item \textbf{Dynamic Steering Mechanism}: We introduce learnable steering vectors that continuously optimize the alignment between time series and textual representations throughout the forward propagation process.
\item \textbf{Parameter-efficient Adaptation}: Our approach only selectively updates layer-specific steering vectors and output model projections, achieving computational efficiency comparable to lightweight models. 
\item \textbf{Few-shot Forecasting Superiority}: Experimental results demonstrate state-of-the-art performance across multiple time series forecasting benchmarks, with an average 31.8\% improvement over baselines.
\end{itemize}

\section{Related Work}
\paragraph{\textbf{Deep Learning-based Time Series Forecasting}} Time series forecasting (TSF) aims to predict future values based on historical observations. Over the past decades, TSF methods have evolved from traditional statistical approaches to machine learning-based and deep learning-based techniques \cite{benidis2022deep, hu2025timefilter, liu2025stochastic, peng2025semantics}. Two key paradigms have primarily driven recent advances in TSF. First, \textbf{Transformer-based methods} leverage self-attention mechanisms to capture long-range dependencies\cite{lu2025timecapsule}, exemplified by Informer’s sparse attention for computational efficiency \cite{zhou2021informer}, Autoformer’s trend-seasonality decomposition \cite{wu2021autoformer}, FEDformer’s hybrid time-frequency processing \cite{zhou2022fedformer}, and so on. In addition, \cite{wang2025pre} employs a two-layer Transformer architecture with three customized pre-training tasks to capture unique statistical features of stock data in the specific financial domain. Beyond standard Transformer adaptations, TimeBridge\cite{liu2024timebridge} has introduces dual attention mechanisms (Integrated/Cointegrated) that differentially handle non-stationarity across time scales, significantly improving forecasting accuracy.
While effective at dealing with complex temporal patterns, these models face generalization challenges across diverse TS domains. On the other hand, \textbf{Time series foundation models} address this via large-scale pretraining \cite{liu2025sundial, shao2025blast}. For example, TimesFM employs patch-based pretraining for zero-shot adaptability across domains \cite{das2024decoder}, and MOMENT proposes ``Time Series Pile'' that enables few-shot proficiency in diverse tasks \cite{goswami2024moment}. MOIRAI innovates its pretraining process with multi-patch projections and any-variate attention for cross-frequency learning \cite{woo2024unified}. Despite the widespread availability of TSFMs fueled by growing interest in their pretraining, effectively leveraging TSFMs with LLMs remains underexplored, limiting the full exploitation of their potential for expressive time series representations.

\paragraph{\textbf{Prompting}} Prompt learning has emerged as a key paradigm to adapt large language models (LLMs) to downstream tasks without modifying model parameters, primarily through task reformulation using textual or vectorized prompts \cite{liu2023pre}. Existing approaches fall into three categories based on representation and optimization. (1) \textbf{Manual hard prompts} employ human-designed templates (e.g., cloze-style "[X] was born in [MASK]" or task-specific prefixes) to guide LLM inference, but suffer from sub-optimality and poor cross-task generalization due to their reliance on expert knowledge \cite{petroni2019language, brown2020language, schick2020exploiting, jiangfstllm}. (2) \textbf{Automated hard prompts} address this by algorithmically generating templates, either through corpus-based pattern mining \cite{jiang2020can} or gradient-guided discrete token optimization \cite{shin2020autoprompt}. But these methods remain constrained by the combinatorial complexity of discrete search spaces. (3) \textbf{Automated soft prompts} overcome limitations mentioned above by operating in continuous embedding spaces. For example, \citet{liu2024gpt} adopts P-tuning to encode prompts via trainable LSTMs, \citet{li2021prefix} employs the Prefix-Tuning to inject learnable prefix vectors into each LLM layer, which enables fine-grained control over LLM behavior. 

\begin{figure}
    \centering
    \includegraphics[width=1.0\linewidth]{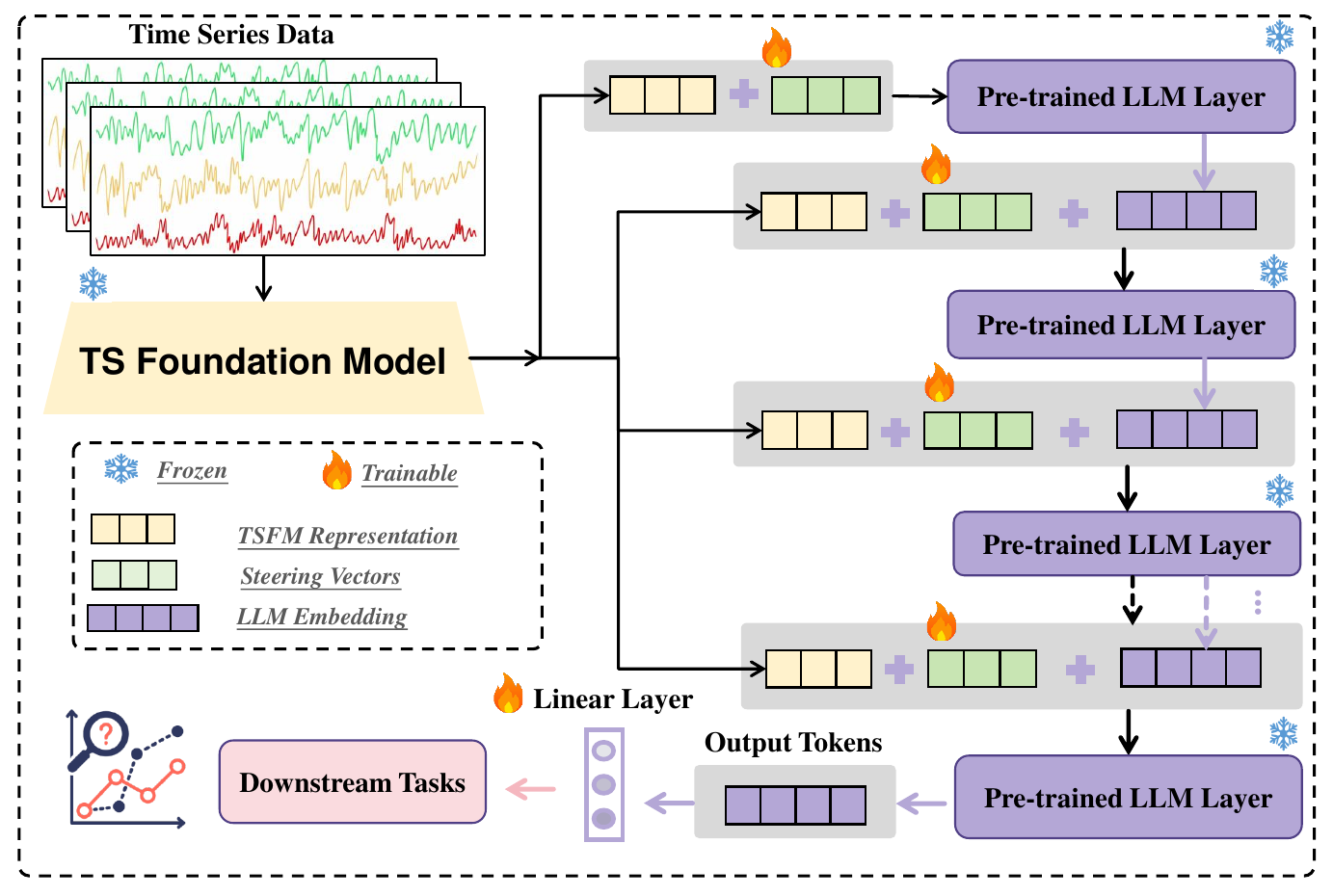}
    \vskip -1.0em
    \caption{The framework of MSEF. On the left part, time series sequences are first dealt with by time series foundation models to generate TSFM representations. On the right part, the LLM textual embeddings (in purple) from the last LLM layer are concatenated with TSFM representations (in yellow) and steering vectors (in green), and then passed to the next LLM layer.
    }
    \label{fig:framework}
    \vspace{-1.0em}
\end{figure}

\section{Methodology}
\subsection{Overview}
We develop a parameter-efficient framework for time series forecasting that adapts pre-trained language models (such as Pythia\cite{biderman2023pythia}) without fine-tuning the backbone model. Specifically, we consider the following problem: Given a historical observation sequence $X_T\in\mathbb{R}^{N\times T}$ consisting of N univariate data streams that span over T time steps, our objective is to allow the language model to accurately understand the time series data and forecast future values $\widehat{X}_H\in\mathbb{R}^{N\times H}$ for H subsequent steps through a few shot adaptation. The optimization goal is to minimize the Mean Squared Error (MSE) between predicted values $\widehat{X}_H$ and ground truth values $X_H$. 

The proposed framework is depicted in Figure \ref{fig:framework}. Specifically, the time series foundation model (such as Moment \cite{goswami2024moment}) is first used to perform deep representation learning on the original TS data to extract hierarchical representations. Then, for each layer of the pretrained LLM, layer-specific steering vectors are used to dynamically align these temporal representations with the language model's textual embedding space. Finally, the adapted representations pass through a lightweight output projection layer (such as a linear layer) to generate forecasting results. It is worth mentioning that only the steering vectors and the prediction layer require training in this framework, while both the language model and the time series encoder remain frozen, ensuring computational efficiency.

\subsection{Multi-layer Steerable Embedding Fusion}
\paragraph{\textbf{Multi-layer TS Fusion}}
We maintain temporal pattern integrity through layer-wise injection of time series representations. Unlike conventional approaches that only process time series embeddings at the input layer, we establish direct feature pathways at each LLM layer. Specifically, the time series foundation model (TSFM) first extracts hierarchical representations from the raw input data, which are then concatenated with trainable steering vectors for injection at each LLM layer. This design creates a continuous temporal representation pathway that preserves the original TS patterns throughout the depth of the LLM and effectively prevents the feature dissipation commonly observed in deeper LLM layers.

\paragraph{\textbf{Dynamic Steering Mechanism}}
With further scrutiny of the TS adaptation issue, the limitations of adding merely a steering vector to the first LLM Layer (input layer) or adding a unified steering vector to each layer are twofold. Firstly, the input layer has limited adjustable parameter capacity due to relatively short vector length, leading to inadequate encoding of complex TS patterns. Secondly, the initial information must pass through dozens of layers of forward propagation to reach the prediction results, where the TS information inserted into the input layer will vanish as the depth of the LLM network increases \cite{liu2021p}. To break this bottleneck, as shown in Figure \ref{fig:framework}, the proposed framework injects differentiated steering vectors into each layer of the LLM, which increases the number of adjustable parameters and makes the vectors have a more direct impact on the model's prediction. Specifically, the trainable dynamic steering vectors and the representations of TS data are first concatenated as prefixes of the output of each LLM layer. Then, they are passed to the next LLM layer as input. On one hand, by continuously injecting high-level representations of TS data at each layer, the LLM is forced to focus on the original TS pattern during processing, avoiding information dilution as network depth increases. 
On the other hand, MSEF could dynamically adjust the expression of TS patterns via updating the steering vectors injected at each layer according to the semantic abstraction captured by the TSFM at different depths.
Most importantly, this design realizes coordinated time series understanding and language processing. For example, LLM can focus on original feature alignment at the bottom layer, pattern recognition at the middle layer, and semantic interpretation at the top layer, thereby forming a hierarchical time series understanding capability. This mechanism not only enhances the model's ability to capture complex temporal dynamics but also, through a continuous optimization process, enables temporal features and language representation to achieve an adaptive optimal balance in the deep network.

\begin{table*}[t!]
\caption{Multivariate time series forecasting under a few-shot setting (10\% train steps). All results are averaged from four different forecasting horizons: H $\in$ \{96, 192, 336, 720\}. The best value for each metric is highlighted in red. A lower value indicates better performance.}
\vskip -1.0em 
\label{tab:10few-shot}
\begin{tabular}{c|cc|cc|cc|cc|cc|cc|cc|cc}
\hline
 &  \multicolumn{2}{c|}{Ours} & \multicolumn{2}{c|}{TimesNet} & \multicolumn{2}{c|}{FEDformer} & \multicolumn{2}{c|}{Autoformer} & \multicolumn{2}{c|}{Stationary} & \multicolumn{2}{c|}{ETSformer} & \multicolumn{2}{c|}{LightTS} & \multicolumn{2}{c}{Informer} \\

& MSE & MAE & MSE & MAE & MSE & MAE & MSE & MAE & MSE & MAE & MSE & MAE & MSE & MAE & MSE & MAE 
\\
\hline
\multirow{1}{*}{ETTh1}
        & {\color[HTML]{FE0300} \textbf{0.451}} & {\color[HTML]{FE0300} \textbf{0.463}} 
            & 0.869 & 0.628  & 0.639 & 0.561  & 0.702  & 0.596         & 0.915 & 0.639  & 1.180 & 0.834  & 1.375  & 0.877
            & 1.199 & 0.809     \\
        \hline
        
\multirow{1}{*}{ETTh2}
        & {\color[HTML]{FE0300} \textbf{0.376}} & {\color[HTML]{FE0300} \textbf{0.411}} 
        & 0.479 & 0.465    & 0.466 & 0.475  & 0.488  & 0.499  
        & 0.462 & 0.455    & 0.894 & 0.713  & 2.655  & 1.160  
        & 3.872 & 1.513     \\
        \hline
        
\multirow{1}{*}{ETTm1}
        & {\color[HTML]{FE0300} \textbf{0.371}} & {\color[HTML]{FE0300} \textbf{0.399}} & 0.677 & 0.537                        & 0.722 & 0.605                        & 0.802          & 0.628         & 0.797          & 0.578         & 0.980         & 0.714         & 0.971        & 0.705        & 1.192         & 0.821        \\
        \hline

\multirow{1}{*}{ETTm2}
        & {\color[HTML]{FE0300} \textbf{0.283}} & {\color[HTML]{FE0300} \textbf{0.337}} & 0.320 & 0.353                        & 0.463 & 0.488                        & 1.342          & 0.930         & 0.332          & 0.366         & 0.447         & 0.487         & 0.987        & 0.756        & 3.370         & 1.440      \\
\hline
\multirow{1}{*}{ECL}
        & {\color[HTML]{FE0300} \textbf{0.236}} & {\color[HTML]{FE0300} \textbf{0.334}} & 0.323 & 0.392                        & 0.346 & 0.427                        & 0.431          & 0.478         & 0.444          & 0.480         & 0.660         & 0.617         & 0.441        & 0.489        & 1.195         & 0.891       \\
\hline
\multirow{1}{*}{Weather}
        & {\color[HTML]{FE0300} \textbf{0.229}} & {\color[HTML]{FE0300} \textbf{0.270}} & 0.279 & 0.301                        & 0.284 & 0.324                        & 0.300          & 0.342         & 0.318          & 0.323         & 0.318         & 0.360         & 0.289        & 0.322        & 0.597         & 0.495       \\
\hline
\multirow{1}{*}{Traffic}
        & {\color[HTML]{FE0300} \textbf{0.542}} & {\color[HTML]{FE0300} \textbf{0.393}} & 0.951 & 0.535                        & 0.663 & 0.425                        & 0.749          & 0.446         & 1.453          & 0.815         & 1.914         & 0.936         & 1.248        & 0.684        & 1.534         & 0.811      \\
\hline

\hline
\end{tabular}
\end{table*}

\section{Experiments}
\subsection{Experimental Settings}
We evaluate the proposed MSEF through few-shot long-term forecasting experiments, where models are trained on only the first 10\% of available time steps to simulate data-scarce scenarios. The benchmark encompasses seven widely adopted datasets: four from the ETT datasets (ETTh1, ETTh2, ETTm1, and ETTm2) \cite{zhou2021informer}, one from the Electricity (ECL)\footnote{UCI. Electricity. https://archive.ics.uci.edu/ml/datasets/ElectricityLoadDiagrams20112014.}, one from the Weather\footnote{Wetterstation. Weather. https://www.bgc-jena.mpg.de/wetter/.} and one from the Traffic\footnote{PeMS. Traffic. http://pems.dot.ca.gov/.}. Those datasets have been extensively adopted for benchmarking long-term forecasting models\cite{wu2022timesnet}. The input sequence length T is set as 512. We test four different prediction horizons $H \in \{96, 192, 336, 720\}$. The model architecture integrates two frozen foundation models: Pythia-1B \cite{biderman2023pythia} with 32 layers for feedforward propagating textual embeddings and the MOMENT-1-large \cite{goswami2024moment} is adopted for generating general time series representations. The evaluation metrics include Mean Squared Error (MSE) and Mean Absolute Error (MAE). 

All experiments are implemented in PyTorch and conducted on 2 NVIDIA A100-40GB GPUs, optimized via Adam with early stopping based on validation loss. Our code is available at \textcolor{magenta}{\url{https://github.com/One1sAll/MSEF}}. 

\subsection{Main Results}
We compare the few-shot multivariate forecasting performance of MSEF to baseline supervised forecasting models, including TimesNet \cite{wu2022timesnet}, FEDformer \cite{zhou2022fedformer}, Autoformer \cite{wu2021autoformer}, Stationary \cite{liu2022non}, ETSformer \cite{woo2022etsformer}, LightTS \cite{zhang2022less}, Informer \cite{zhou2021informer}. All the above baselines are trained on the 10\% train split of each benchmark dataset. Note that MSEF achieves the best result in 69 out of 70 comparisons. Due to the page limit, we only list the average performance of four tested prediction horizons (H$\in$ \{96, 192, 336, 720\}). As illustrated in Table \ref{tab:10few-shot}, MSEF outperforms all baselines in all cases.
Specifically, the proposed MSEF outperforms TimesNet by 31.8\%. Compared with the task-specific Transformer-based model FEDformer, MSEF achieves an average MSE reduction of 31.1\%. As for methods related to other backbones, such as LightTS, MSEF achieves a performance improvement as significant as 59.5\%.

\begin{figure}
    \centering
    \includegraphics[width=1.0\linewidth]{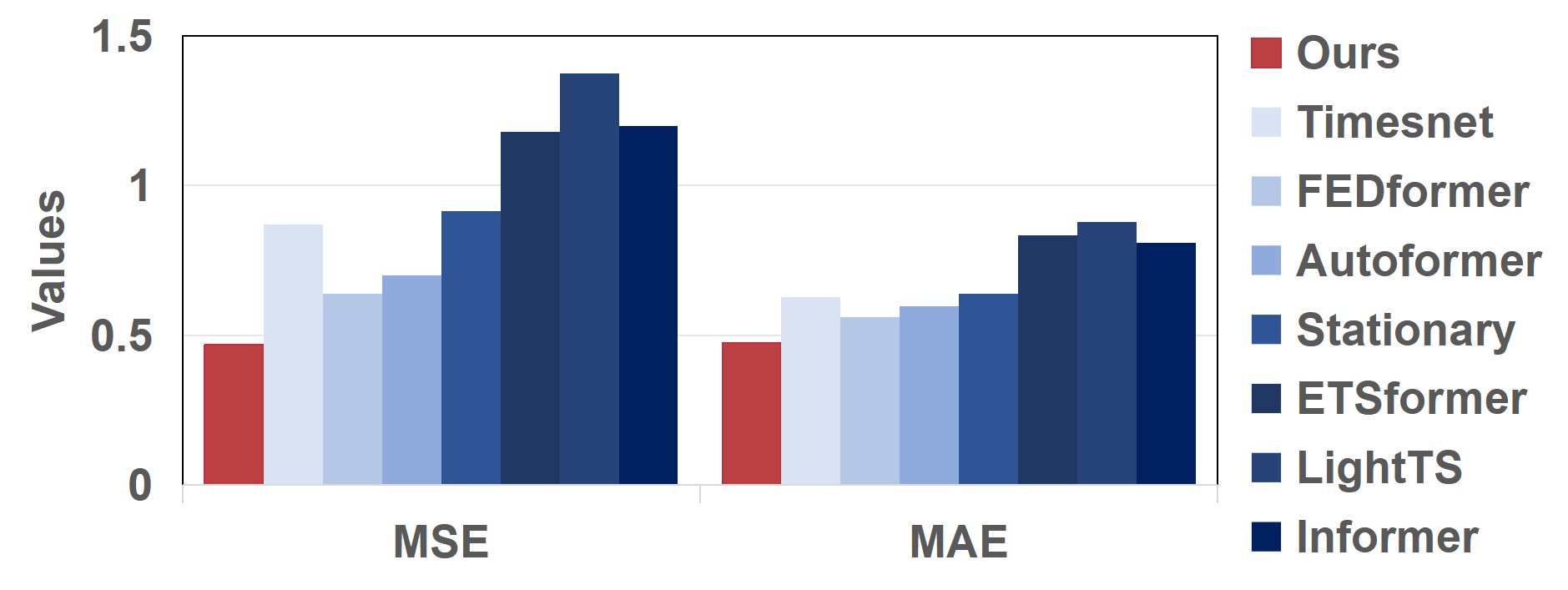}
    \vskip -1.0em
    \caption{Comparison of average performance of various methods on ETTh1 dataset. }
    \label{fig:ETTh1}
    \vspace{-1.0em}
\end{figure}

Additionally, we visualize the average performance comparison of each time series forecasting method on ETTh1 in Figure \ref{fig:ETTh1}. MSEF achieves the best results in both MSE (0.468) and MAE (0.474), with its MSE significantly outperforming the suboptimal model (FEDformer, 0.639), while LightTS performs worst in both metrics, standing in sharp contrast to our model's excellent performance.

\subsection{Ablation Study}
To analyze the contribution of our method, we conduct an ablation study on the ETTh1 and Weather datasets. As Table \ref{tab:ablation-study} shows, first, we convert the time series data into raw text descriptions and directly pass them to LLM (we call this test the ``Plain LLM-based'' study, namely testing without both TSFM representations and steering vectors). It can be seen that due to the large sequence length of TS data and its semantic sparsity, LLM struggles in making accurate predictions based on the plain numerical-to-text/token conversion of both datasets. Secondly, we directly concatenate the TSFM representations at each layer of LLM (w/o steering vectors). 
It can be seen that LLM with TSFM representations as injected information outperforms the plain one. However, as shown in the table, LLM with both TSFM representations and steering vectors (ours) shows the best performance. 


\begin{table}[]
\caption{Ablation study. Plain LLM-based testing with LLM model w/o both TSFM representations and steering vectors. For the testing w/o steering vectors, TSFM representations are injected into the LLM in a layer-specific manner.}
\small
\vskip -1.0em
\label{tab:ablation-study}
\begin{tabular}{cc|cc|cc|cc}
\hline
 &  & \multicolumn{2}{c|}{Plain LLM-based}                                        & \multicolumn{2}{c|}{w/o steering vectors} & \multicolumn{2}{c}{Ours} \\
 &  & MSE                                   & MAE                                   & MSE                & MAE                & MSE           & MAE          \\
\hline
                        & 96   & 1.676         & 1.875                 & 0.425        & 0.441   
                        & {\color[HTML]{FE0300} \textbf{0.413}} 
                        & {\color[HTML]{FE0300} \textbf{0.435}}                   \\
                        & 192  & 1.673         & 1.874
                        & 0.496         & 0.489               
                        & {\color[HTML]{FE0300} \textbf{0.441}} 
                        & {\color[HTML]{FE0300} \textbf{0.452}}\\
                        & 336  & 1.664         & 1.872
                        & 0.497         & 0.495                  
                        & {\color[HTML]{FE0300} \textbf{0.458}} 
                        & {\color[HTML]{FE0300} \textbf{0.467}}    \\
                        & 720  & 1.491         & 1.855
                        & 0.561   & 0.530                 
                        & {\color[HTML]{FE0300} \textbf{0.494}} 
                        & {\color[HTML]{FE0300} \textbf{0.499}}\\
\multirow{-5}{*}{\rotatebox{90}{ETTh1}} & avg                  & 1.626    & 1.869   & 0.495   & 0.489      
& {\color[HTML]{FE0300} \textbf{0.451}} 
& {\color[HTML]{FE0300} \textbf{0.463}} \\

\hline
                        & 96   & 0.574         & 0.470                 & 0.161        & 0.221   
                        & {\color[HTML]{FE0300} \textbf{0.153}} 
                        & {\color[HTML]{FE0300} \textbf{0.209}}                   \\
                        & 192  & 0.573         & 0.469
                        & 0.210         & 0.264               
                        & {\color[HTML]{FE0300} \textbf{0.197}} 
                        & {\color[HTML]{FE0300} \textbf{0.248}}\\
                        & 336  & 0.573         & 0.470
                        & 0.260         & 0.302                  
                        & {\color[HTML]{FE0300} \textbf{0.247}} 
                        & {\color[HTML]{FE0300} \textbf{0.285}}    \\
                        & 720  & 0.591         & 0.511
                        & 0.318   & 0.341                 
                        & {\color[HTML]{FE0300} \textbf{0.317}} 
                        & {\color[HTML]{FE0300} \textbf{0.337}}\\
\multirow{-5}{*}{\rotatebox{90}{Weather}} & avg                  & 0.578    & 0.480   & 0.237   & 0.282      
& {\color[HTML]{FE0300} \textbf{0.229}} 
& {\color[HTML]{FE0300} \textbf{0.270}} \\
\hline
\end{tabular}
\vspace{-1.0em}
\end{table}

\begin{figure}
    \centering
    \includegraphics[width=1\linewidth]{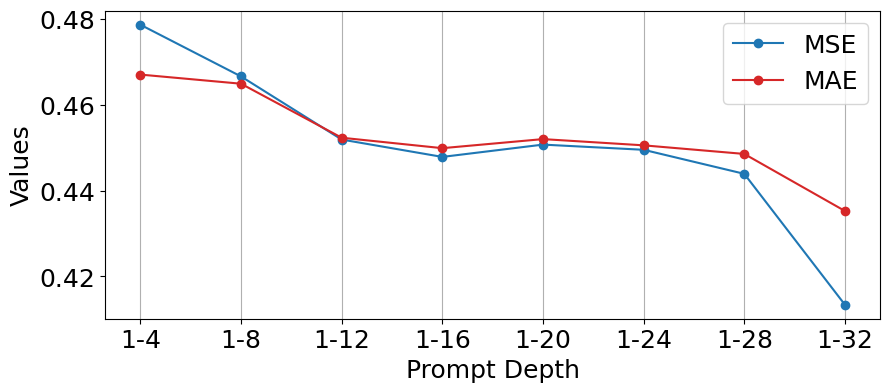}
    \vskip -1.0em
    \caption{ Ablation study on steering vector's depth (on ETTh1). “[x-y]" refers to the layer interval added with continuous vectors (e.g., “1-4” means we add steering vectors to LLM layers from 1 to 4). }
    \label{fig:prompt depth}
    \vspace{-1.0em}
\end{figure}

\section{Conclusion}
In this paper, we proposed the \textbf{Multi-layer Steerable Embedding Fusion (MSEF)} framework, which is a novel approach for adapting large language models to few-shot time series forecasting without fine-tuning the backbone model. By dynamically generating and injecting layer-specific steering vectors, MSEF effectively bridges the gap between time series data and the pattern extrapolation capabilities of large language models (LLMs) pretrained on large-scale textual corpora. This mechanism enables the model to capture complex temporal patterns while preserving essential information throughout the feedforward propagation across LLM layers, ultimately leading to improved forecasting performance. Experiments on seven benchmark datasets demonstrate significant outperformance in few-shot settings over existing supervised baselines.

\section*{GenAI Usage Disclosure}
In accordance with related policies, we disclose that generative AI tools were used solely for language editing purposes during the preparation of this manuscript. Specifically, AI-assisted tools (Deepseek, Grammarly) were only employed to check and correct grammatical errors and typos, improve sentence clarity, and ensure consistent academic writing style. All AI-processed text was carefully reviewed and approved by the authors, with the research content, methodology, results, and conclusions remaining entirely human-generated. No generative AI was used in any aspect of the research design, data analysis, code development, or technical interpretation of results. The final manuscript represents the original intellectual contributions of the authors alone.

\bibliographystyle{ACM-Reference-Format}
\balance
\bibliography{ref}

\appendix

\end{document}